\documentclass{article}

\usepackage{times}
\usepackage{graphicx} 

\usepackage{natbib}

\usepackage{algorithm}
\usepackage{algorithmic}

\usepackage{hyperref}

\usepackage{booktabs}
\usepackage{color}
\usepackage{placeins}
\usepackage{amsmath}
\usepackage{amsthm}
\usepackage{amssymb}
\usepackage{caption}
\usepackage{subcaption}
\usepackage{tikz}
\usepackage{comment}

\newtheorem{prop}{Proposition}
\newtheorem{conjecture}{Conjecture}
\newtheorem{definition}{Definition}

\newcommand\numberthis{\refstepcounter{equation}\tag{\theequation}}

\newcommand{\stefano}[1]{}
\newcommand{\shengjia}[1]{}

\def\E{\mathbb{E}}
\usepackage[accepted]{icml2017}

\icmltitlerunning{Towards a Deeper Understanding of Variational Autoencoding Models}

\begin{document}

\twocolumn[
\icmltitle{Towards a Deeper Understanding of Variational Autoencoding Models}

\begin{icmlauthorlist}
\icmlauthor{Shengjia Zhao}{stanford}
\icmlauthor{Jiaming Song}{stanford}
\icmlauthor{Stefano Ermon}{stanford}
\end{icmlauthorlist}

\icmlaffiliation{stanford}{Stanford University}

\icmlcorrespondingauthor{Shengjia Zhao}{zhaosj12@stanford.edu}
\icmlcorrespondingauthor{Jiaming Song}{tsong@stanford.edu}
\icmlcorrespondingauthor{Stefano Ermon}{ermon@stanford.edu}
\icmlkeywords{variational, VAE, generative models}

\vskip 0.3in
]
\printAffiliationsAndNotice{}

\begin{abstract}
We propose a new family of optimization criteria for variational auto-encoding models, generalizing the standard evidence lower bound. We provide conditions under which they recover the data distribution and learn latent features, and formally show that common issues such as blurry samples and uninformative latent features arise when these conditions are not met. Based on these new insights, we propose a new sequential VAE model that can generate sharp samples on the LSUN image dataset based on pixel-wise reconstruction loss, and propose an optimization criterion that encourages unsupervised learning of informative latent features.
\end{abstract}

\section{Introduction}


Generative models have made remarkable progress in recent years. Existing techniques can model fairly complex datasets, including natural images and speech~\citep{deconvolutional_gan2015, wgan2017, pixel_rnn2015, conditional_pixelcnn2016, infusion_training2016, vae_autoregressive_flow2016, pixel_vae2016, generate_text2015, recurrent_vae_sequence2015}. 
Latent variable models, such as variational autoencoders (VAE), are among the most successful ones. 
\citep{autoencoding_variational_bayes2013, variational_dbn_stochastic_bp2014, vae_autoregressive_flow2016, ladder_variational_network2015, matnet_variational_network2016, pixel_vae2016}. 
These generative models are very flexible, but
the resulting marginal likelihood 
is intractable to compute and optimize.
Models are thus learned by optimizing a tractable "evidence lower bound", obtained using a tunable inference distribution.

Despite the empirical success, 
existing VAE models are unable to accurately model complex, large scale image datasets such as LSUN and Imagenet, unless other approaches such as adversarial training or supervised features are employed \citep{generative_network_perceptual_metric2016, autoencoder_with_gan_metric2015, discriminative_reg_for_vae2016}. 
When applied to complex datasets of natural images, VAE models tend to produce unrealistic, blurry samples
\citep{generative_network_perceptual_metric2016}. The sample quality can be improved with a more expressive generative model, however, 
this leads to
a tendency to ignore the latent variables, thus 
hindering the unsupervised feature learning goal
\citep{lossy_vae2016}. 

In this paper, we propose a new derivation of VAEs which is not based on a variational Bayes approach.
We propose a new and more general optimization criterion that is not always a lower bound on the marginal likelihood, but it is guaranteed to learn the data distribution under suitable conditions. For a particular choice of regularization, our approach becomes the regular VAE \citep{autoencoding_variational_bayes2013}.
This new derivation gives us insights into the properties of VAE models. In particular, we are able to formally explain some common failure modes of VAEs, and propose novel methods to alleviate these issues.

In Section 3 we provide a formal explanation for why VAEs generate blurry samples when trained on complex natural images. We show that under some conditions, blurry samples are not caused by the use of a maximum likelihood approach as previously thought, but rather they are caused by an inappropriate choice for the inference distribution.
We specifically target this problem by proposing a sequential VAE model, where we gradually augment the the expressiveness of the inference distribution using a process 
inspired by
the recent infusion training process\citep{infusion_training2016}.
As a result, we are able to generate sharp samples on the LSUN bedroom dataset, even using $2$-norm reconstruction loss in pixel space.

In Section 4 we propose a new explanation of the VAE tendency to ignore the latent code. We show that this problem is specific to the original VAE objective function \citep{autoencoding_variational_bayes2013} and does not apply to the more general family of VAE models we propose. We show experimentally that using our more general framework, we achieve comparable sample quality as the original VAE, 
while at the same time learning meaningful features through the latent code,
even when the decoder is a powerful PixelCNN that can by itself model data \citep{pixel_rnn2015, conditional_pixelcnn2016}. 

\section{A Novel Derivation for VAE}
\subsection{Training Latent Variable Models}
Let $p_{data}(x)$ be the true underlying data distribution defined over $x \in \mathcal{X}$, and $\mathcal{D}$ be a dataset of i.i.d. samples from $p_{data}(x)$. 
In the context of unsupervised learning, we are often interested in learning features and representations directly from unlabeled data. A common approach for inducing features is to consider a joint probability distribution $p(x,z)$ over $(x,z)$, where $x \in \mathcal{X}$ is in the observed data space and $z \in \mathcal{Z}$ is a {\em latent code} or {\em feature space}. The distribution $p(x,z)$ is specified with a prior $p(z)$ and a conditional $p(x|z)$. The prior $p(z)$ is often chosen to be relatively simple -- the hope is that the interactions between high level features are disentangled, and can be well approximated with a Gaussian or  uniform distribution. The complexity of $p_{data}(x)$ is instead captured by the conditional distribution $p(x|z)$. For the analysis in this paper, it will be convenient to specify $p(x|z)$ using two components:

\begin{enumerate}
\item A family of probability distributions $\mathcal{P}$ over $\mathcal{X}$. We require that the set $\mathcal{P}$ is parametric, which means that it can be indexed by a set $\Lambda$ in finite dimensional real space $\Lambda \subset \mathbb{R}^D$. We furthermore require that for every $\lambda \in \Lambda$, the corresponding element $\mathcal{P}_\lambda$ has well-defined and tractable log likelihood derivative $\nabla_\lambda \log \mathcal{P}_\lambda(x)$ for any $x \in \mathcal{X}$. 
\item A mapping $f_\theta: \mathcal{Z} \to \Lambda$ parameterized by $\theta$ with well defined and tractable derivatives $\nabla_\theta f_\theta(z)$ for all $z \in \mathcal{Z}$. We also denote the family of all possible mappings defined by our model as $\mathcal{F}=\{f_\theta, \theta \in \Theta\}$. 
\end{enumerate}

Given $\mathcal{P}$ and $\mathcal{F}$, we define a family of models
\[ p_\theta(x, z) = p_\theta(x|z) p(z) = \mathcal{P}_{f_\theta(z)}(x)p(z) \]
indexed by $\theta \in \Theta$. Note that $ \mathcal{P}_{f_\theta(z)}$ plays the role of the conditional distribution $p_\theta(\cdot \mid z)$ for any given $z \in \mathcal{Z}$. For brevity, we will use $p_\theta(x|z)$ to indicate $\mathcal{P}_{f_\theta(z)}(x)$. Note that the resulting marginal likelihood $p_\theta(x)$ is a mixture of distributions in $\mathcal{P}$
\[ p_\theta(x) = \int_z p(z) p_\theta(x|z) dz = \E_{p(z)}[p_\theta(x|z)] \]

While the specification of $p_\theta(\cdot | z)$ using $\mathcal{P}$ and $f_\theta$ is fully general, our definition imposes some (mild) tractability restrictions on $\mathcal{P}$ and $f_\theta$ to allow for efficient learning. Specifically, the models we consider are those where 
$\nabla_\theta p_\theta(x|z)$ can be tractably computed using the chain rule: 
\[ \nabla_\theta p_\theta(x|z) = \frac{\partial \log \mathcal{P}_{\lambda}(x)}{\partial \lambda}^T \frac{\partial \lambda}{\partial \theta} = \frac{\partial \log \mathcal{P}_{\lambda}(x)}{\partial \lambda}^T \frac{\partial f_\theta(z)}{\partial \theta} \] 
This class of models encompasses many recent approaches~\citep{autoencoding_variational_bayes2013, vae_autoregressive_flow2016, conditional_pixelcnn2016, pixel_vae2016}, where $\mathcal{P}$ is often Gaussian or a recurrent density estimator, and $\mathcal{F}$ is a deep neural network.

We consider a maximum likelihood based approach to learn the parameters, where the goal is to maximize the marginal likelihood of the data
\begin{align*}
\max_\theta &\E_{p_{data}(x)}[\log p_\theta(x)] \\ 
&= \max_{\theta} \E_{p_{data}(x)} \left[\log \E_{p(z)}[p_\theta(x|z)] \right] \numberthis \label{equ:maximum_likelihood}
\end{align*}
where the expectation over $p_{data}(x)$ is approximated using a sample average over the training data $\mathcal{D}$.

To actually optimize over the above criteria we require that for any $x$, the derivative over model parameters of $\nabla_\theta \log p_\theta(x)$ be easy to compute or estimate. However tractability of $\nabla_\theta p_\theta(x|z)$ do not imply tractability of $\nabla_\theta p_\theta(x)$, and if we directly take the derivative we get
\begin{align*}
\nabla_\theta \log p_\theta(x) &= p_\theta(x)^{-1} \nabla_\theta E_{p(z)}[p_\theta(x|z)] \\ 
&= p_\theta(x)^{-1} E_{p(z)}[\nabla_\theta p_\theta(x|z)] \numberthis \label{equ:intractable_model}
\end{align*}
Evaluating $p_\theta(x)^{-1}$ involves the computation of a high dimensional integral. Even though this integral can be approximated by sampling 
\[ p_\theta(x) = E_{p(z)}[p_\theta(x|z)] \approx \frac{1}{n} \sum_{z_i, \cdots, z_n \sim p(z)} p_\theta(x|z_i) \]
this costly approximation has to be performed for every sample $x \sim p_{data}(x)$.




\subsection{A Naive Variational Lower Bound}
To gain some intuition, we consider a simple attempt to make Eq.(\ref{equ:maximum_likelihood}) easier to optimize. By Jensen's inequality we can obtain a lower bound
\begin{align*}
\log p_\theta(x) &= \log E_{p(z)}[p_\theta(x|z)] \\
&\geq E_{p(z)}[\log p_\theta(x|z)]  \numberthis \label{equ:incorrect_model}
\end{align*}
The gradient of this lower bound can be computed more efficiently, as it does not involve the intractable estimation of $p_\theta(x)$
\[ \nabla_\theta E_{p(z)}[\log p_\theta(x|z)] = E_{p(z)}[\nabla_\theta \log p_\theta(x|z)] \]
The hope is that maximizing this lower bound will also increase the original log-likelihood $\log p_\theta(x)$. However, this simple lower bound is not suitable. 
We can rewrite this lower bound as
\begin{align*}
E_{p_{data}(x)} & E_{p(z)}[\log p_\theta(x|z)] \\ 
&=  E_{p(z)} \left[ E_{p_{data}(x)}[\log p_\theta(x|z)] \right] \numberthis \label{equ:alternative_form} \\
\end{align*}
No matter what prior $p(z)$ we choose, this criteria is maximized if for each $z \in \mathcal{Z}$, $E_{p_{data}(x)}[\log p_\theta(x|z)]$ is maximized. However, recall that $\forall z \in \mathcal{Z}, \theta \in \Theta$, $p_\theta(x|z) \in \mathcal{P}$. As a result there is an optimal member $p^* \in \mathcal{P}$ \textbf{independent} of $z$ or $\theta$ that maximizes this term. 
\[ p^* \in \arg\max_{p \in \mathcal{P}} \E_{p_{data}(x)}[\log p(x)]] \]
This means that regardless of $z$, if we always choose $p_\theta(\cdot|z) = p^*$, or equivalently an $f_\theta$ which maps all $z$ to the parameter of $p^*$, then Eq.(\ref{equ:alternative_form}) is maximized. For example, if $\mathcal{P}$ is family of Gaussians, we will not learn a mixture of Gaussians, but rather the {\em single} best Gaussian fit to $p_{data}(x)$.
Optimizing this lower bound is easy, but undermines our very purpose of learning meaningful latent features. 
 

\subsection{Using Discrimination to Avoid Trivial Solution}

The key problem demonstrated in Eq.(\ref{equ:alternative_form}) is that for any $z$, we are fitting the same $p_{data}(x)$ with a member of $\mathcal{P}$.
However, if for every $z$ we fit a different distribution, then we will no longer be limited to this trivial solution. 

Suppose we are given a \textbf{fixed} inference distribution $q(z|x)$, which maps (probabilistically) inputs $x$ to features $z$. Even though our goal is unsupervised feature learning, we initially assume the features are given to us. It is much easier to understand the dynamics of the model when we take $q$ to be fixed. Then we generalize our understanding to learned $q$ in the next section. 
\begin{definition}
We define a joint distribution $q(x, z) = p_{data}(x)q(z|x)$, a marginal $q(z) = \int_z q(x, z) dz$, and a posterior $q(x|z) = q(z|x)p_{data}(x) / q(z)$.  
\end{definition}
In contrast with a standard variational Bayes approaches, for now we do {\em not} treat $q$ as a variational approximation to the posterior of some generative model $p$. Instead we simply take $q$ to be any distribution that probabilistically maps $x$ to features $z$. For example, $q$ can be a classifier that detects object categories.

We define a new optimization criteria where for each $z$ we use a member of $\mathcal{P}$ to fit a different $q(x|z)$ rather than the entire $p_{data}$. 
\begin{align*}
\mathcal{L} = \E_{q(z)} \left[ \E_{q(x|z)} [\log p_\theta(x|z)] \right] \numberthis \label{equ:model_with_q}
\end{align*}
Comparing with Eq.(\ref{equ:alternative_form}), there is a key difference. As before, no matter what $q(z)$ we choose, this new criteria is maximized when for each $z \in \mathcal{Z}$, $E_{q(x|z)} [\log p_\theta(x|z)]$ is maximized, or equivalently $KL(q(x|z)||p_\theta(x|z))$ is minimized, {\em as a function of } $\theta$ (because $q$ is fixed). However, in contrast with Eq.(\ref{equ:alternative_form}) we approximate a different $q(x|z)$ for each $z$ , rather than finding the single best distribution in $\mathcal{P}$ to fit the {\em entire} data distribution $p_{data}(x)$. 

While this is no longer a (lower) bound on the marginal likelihood as in Eq. (\ref{equ:incorrect_model}), we now show that under some conditions this criterion is suitable for learning.

\textbf{1) Tractable stochastic gradient:} This criteria admits a tractable stochastic gradient estimator because
\begin{align*}
\nabla_\theta \mathcal{L} &= \nabla_\theta E_{q(x, z)} [\log p_\theta(x|z)]  \\ 
&=  E_{q(x, z)} [\nabla_\theta \log p_\theta(x|z)] 
\end{align*}
As before it can be efficiently optimized using mini-batch stochastic gradient descent. 

\textbf{2) Utilization of Latent Code:} As our intuition that motivated the design of this  objective (\ref{equ:model_with_q}) points out, this criteria incentivizes the use of latent code that gives ``discriminative power'' over $x$, which we formally demonstrate in the following proposition
\begin{prop}
\label{prop:optimal_solution}
Let $\theta^*$ be the global optimum of $\mathcal{L}$ defined in (\ref{equ:model_with_q}), and $f_{\theta^*}$ the corresponding optimal mapping. If $\mathcal{F}$ has sufficient capacity, then for every $z \in \mathcal{Z}$
\[ \mathcal{P}_{f_{\theta^*}(z)} \in \arg\max_{p \in \mathcal{P}} E_{q(x|z)}[\log p(x)] \]
\end{prop}

\begin{figure*}
\begin{subfigure}[t]{0.5\textwidth}
\centering
\includegraphics[height=0.4\linewidth]{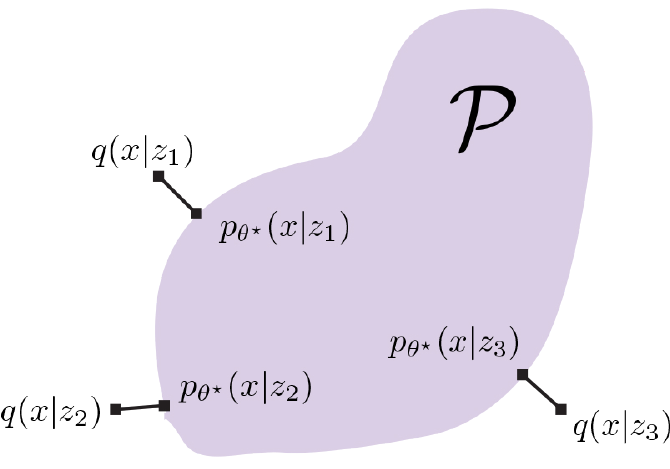}
\end{subfigure}
\begin{subfigure}[t]{0.5\textwidth}
\centering
\includegraphics[height=0.4\linewidth]{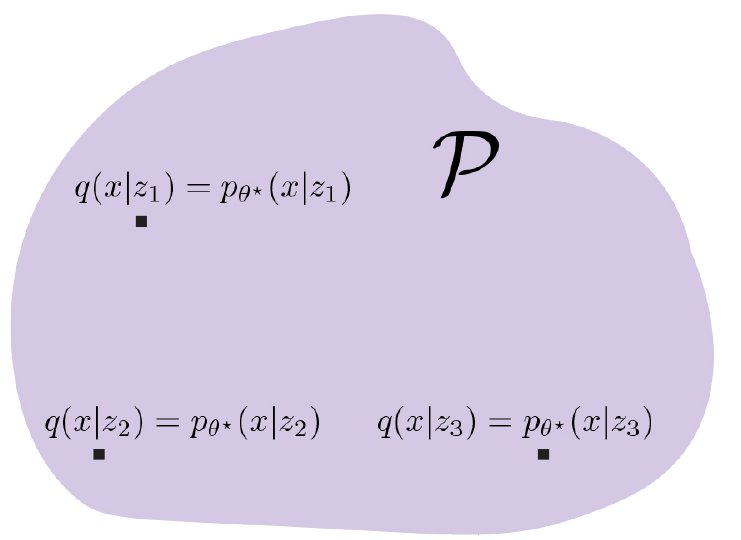}
\end{subfigure}
\caption{Illustration of variational approximation of $q(x|z)$ by $\mathcal{P}$. Left: for each $z \in \mathcal{Z}$ we use the optimal member of $\mathcal{P}$ to approximate $q(x|z)$. Right: this approximation requires $\mathcal{P}$ to be large enough so that it covers the true posterior $q(x|z)$ for any $z$.}
\label{fig:variation_illustration}
\end{figure*}

This is illustrated in Figure~\ref{fig:variation_illustration}. When $\mathcal{F}$ has sufficient representation capacity, we are using $\mathcal{P}$ to variationally approximate $q(x|z)$ for each $z \in \mathcal{Z}$ respectively.

For example, suppose $\mathcal{X}$ are images and $q(z|x)$ is an image classifier over $K$ classes. 
Then $q(x|z=k)$ can be thought as the appeareance distribution of objects belonging to class $k$. According to Proposition \ref{prop:optimal_solution}, the optimal generative model based on this inference distribution $q(z|x)$  and objective (\ref{equ:model_with_q}) will select a member of $\mathcal{P}$ to approximate the distribution over images {\em separately for each category}. The optimal $f_{\theta^*}$ will map each object category to this optimal category-specific approximation, assuming it has enough capacity. 

On the other hand, if a feature does not carry discriminative information about $x$, i.e. $q(x|z_1) = q(x|z_2)$, then the optimal $f_{\theta^*}$ will have no motivation to map them to different members of $\mathcal{P}$. $\mathcal{L}$ is already maximized if both are mapped to the same optimal $p^*$ of $\mathcal{P}$ that approximates $q(x|z_1)$ or $q(x|z_2)$. We will return to this point below when we discuss learning $q(z|x)$. 


\textbf{3) Estimation of $p_{data}$:} We further show that under suitable conditions this new learning criterion is consistent with our original goal of modeling $p_{data}(x)$.

\begin{prop}
\label{prop:condition}
Let $\theta^*$ be the global optimum of $\mathcal{L}$ in Equation (\ref{equ:model_with_q}) for a sufficiently large $\mathcal{F}$. If $\mathcal{P}$ is sufficiently large so that
\[ \forall z \in \mathcal{Z}, q(x|z) \in \mathcal{P} \]

then the joint distribution $q(z)p_{\theta^*}(x|z)$ has marginal $p_{data}(x)$, and the Gibbs chain
\begin{align*}
z^{(t)} &\sim q(z|x^{(t)}) \\
x^{(t+1)} &\sim p_{\theta^*}(x|z^{(t)}) \numberthis \label{equ:markov_sampling}
\end{align*}
converges to $p_{data}(x)$ if it is ergodic.
\end{prop}

This condition is illustrated in Figure~\ref{fig:variation_illustration}. Intuitively, this means that if $\mathcal{P}$ is sufficiently large and can exactly represent the posterior $q(x|z)$, then our approach will learn $p_{data}$. Note however that it is $q(x,z)=q(z)p_{\theta^*}(x|z)$ that has marginal $p_{data}(x)$, and {\em not} the original generative model $p(x,z) =p(z)p_{\theta^*}(x|z)$. Nevertheless, if the conditions are met we will have learned all that is needed to sample or draw inferences from $p_{data}(x)$, for example, by the Gibbs chain defined in Proposition~\ref{prop:condition}.

The significance of this result is that $q(z|x)$ can be \textbf{any} feature detector. As long as its posterior can be represented by $\mathcal{P}$, we will learn $p_{data}$ by optimizing (\ref{equ:model_with_q}) with a sufficiently expressive family $\mathcal{F}$.  We will show that this leads to a important class of models in the next section.  

One drawback of the proposed approach is that we cannot tractably sample from $p_{data}(x)$ with ancestral sampling, because the marginal of the original generative model $p(x,z) =p(z)p_{\theta^*}(x|z)$ will not match the data distribution $p_{data}(x)$ in general.
To do ancestral sampling on $p(z) p_\theta(x|z)$, we need an additional condition
\begin{prop}
\label{prop:marginal_condition}
If all conditions in Proposition~\ref{prop:condition} hold, and we further have
\[ \forall z \in \mathcal{Z}, p(z) = q(z) \]
then the original generative model $p(z)p_{\theta^*}(x|z)$ has marginal $p_{data}(x)$. 
\end{prop}
Enforcing this extra condition would restrict us to use only inference distributions $q(z|x)$ whose marginal $q(z)$ matches the prior $p(z)$ specified by the generative model. This is the first time we are placing constraints on $q$. Such constraints generally require {\em joint} learning of $p_{\theta}(x|z)$ and $q$, which we will discuss next. 

\subsection{Learning an Inference Distribution}
In the previous section we assumed that the inference distribution $q$ was fixed, and already given to us. However, in unsupervised learning feature detectors are generally not given a-priori, and are the main purpose of learning itself. In this section we discuss learning a $q$ so that conditions in Proposition~\ref{prop:condition} (and potentially \ref{prop:marginal_condition}) are satisfied. 

Suppose $q$ is also a parameterized distribution with parameters $\phi \in \Phi$, and we denote it as $q_\phi$. As required in VAE models in general \citep{autoencoding_variational_bayes2013} we require $q$ to be reparameterizable so that $\nabla_\phi E_{q_\phi(x, z)}[f(x, z)]$ can also be effectively approximated by stochastic gradients.


\textbf{1) Unregularized VAE:} According to Proposition~\ref{prop:condition}, if do not require tractable sampling from $p_{data}(x)$, then we can simply jointly optimize $\phi$ and $\theta$ under the original objective in in Equation (\ref{equ:model_with_q}). 
\[ 
\max_{\phi, \theta} E_{q_\phi(x, z)}[\log p_\theta(x|z)] \]
Intuitively, we are not only using $\mathcal{P}$ to approximate $q_\phi(x|z)$ for each $z \in \mathcal{Z}$, but we are also learning a $q_\phi$ such that its posterior is simple enough to be representable by $\mathcal{P}$. Successful training under this criterion allows us to model $p_{data}(x)$ by a Gibbs Markov chain (\ref{equ:markov_sampling}). We refer to this model as \textbf{unregularized VAE}. These models do not allow direct (tractable) sampling, but they have desirable properties that we will discuss and evaluate experimentally in  Section~\ref{sec:information_preference}.

\textbf{2) VAE with Regularization}. If we would like to directly (and tractably) sample from $p(z)p_\theta(x|z)$ and have marginal $p_{data}(x)$, then we also need to have $p(z) = q_\phi(z)$. A general way to enforce this condition is by a regularization that penalizes deviation of $q_\phi(z)$ from $p(z)$ with some $R(q_\phi) > 0$, and $R(q_\phi) = 0$ if and only if $p(z) = q_\phi(z)$. The optimization criteria becomes
\begin{align*}
\mathcal{L}_{VAE} = E_{q_\phi(x, z)} [\log p_\theta(x|z)] - R(q_\phi) \numberthis \label{equ:vae_family}
\end{align*}
This gives us a new family of variational auto-encoding models. In particular when $R(q_\phi) = E_{p_{data}(x)} [KL(q_\phi(z|x)||p(z))]$ we get the well known ELBO training criteria \citep{autoencoding_variational_bayes2013}
\begin{align*}
\mathcal{L}_{ELBO} &= E_{p_{data}}[-KL(q_\phi(z|x)||p(z))] \\ 
& \qquad \qquad + E_{q_\phi(x, z)}[\log p_\theta(x|z)] \numberthis \label{equ:elbo_criteria}
\end{align*}
However, ELBO is only one of many possibilities. ELBO has an additional advantages in that it gives us a lower bound for the log-likelihood $\log p_\theta(x)$
\[ \log p_\theta(x) \geq -KL(q_\phi(z|x)||p(z)) + E_{q_\phi(z|x)}[\log p_\theta(x|z)] \]
However the ELBO also has significant disadvantages that we shall discuss in Section~\ref{sec:information_preference}. 

To summarize, our new derivation provides two insights, which lay the foundation for all discussions in the rest of this paper:

\textbf{1) Jointly optimizing $q_\phi(z|x)$ and $p_\theta(x|z)$ with a sufficiently flexible family $\mathcal{F}$ attempts to learn a feature detector such that its posterior $q_\phi(x|z)$ is representable by $\mathcal{P}$.} We will explain in Section~\ref{sec:weak_p_vs_q} that many existing problems with VAEs arise because of the inability of $\mathcal{P}$ to approximate the posterior of $q$. We will also propose a solution that targets this problem.


\textbf{2) We can use any regularization $R(q_\phi)$ that encourages $q_\phi(z)$ to be close to $p(z)$, or no regularization at all if we do not need ancestral sampling}. This will be the central topic of Section~\ref{sec:information_preference}.

\section{Simple $\mathcal{P}$ Requires Discriminative $q$}
\label{sec:weak_p_vs_q}
By our previous analysis, the posterior of $q_\phi$ should be representable by $\mathcal{P}$. For many existing models, although $f_\theta$ is complex, $\mathcal{P}$ is often chosen to be simple, such as the Gaussian family \citep{autoencoding_variational_bayes2013, variational_dbn_stochastic_bp2014, importance_weighted_autoencoders2015}, or a fully factorized discrete distribution \citep{autoencoding_variational_bayes2013}. Proposition 2 requires $q_\phi$ to have a posterior $q_\phi(x|z)$ (the conditional data distribution corresponding to feature $z$) which is also simple. 
We claim that several existing problems of VAE models occur when this condition is not met. 

\subsection{Limitations of Gaussian conditionals $\mathcal{P}$}
\label{sec:gaussian_limitation}
One commonly observed failure with auto-encoding models is the generation of blurry or fuzzy samples. This effect is commonly associated with AE/VAE models that use the L2 loss \citep{generative_network_perceptual_metric2016}. In this setting, we map from data $x$ to latent code $z$ through a encoder $q_\phi(z|x)$, and then reconstruct through a decoder $\hat{x} = g_\theta(z)$. Loss is evaluated by 2-norm of reconstruction error 
\begin{align*} 
\mathcal{L}_{Recon} = E_{p_{data}(x)}E_{q_\phi(z|x)}[||g_\theta(z) - x||_2^2] - R(q_\phi) \numberthis \label{equ:reconstruction_loss}
\end{align*}
where $R(q_\phi)$ is some regularization on $q_\phi$. Note that if we define the distribution $p_\theta(x|z) = \mathcal{N}(g_\theta(z), I/2)$, then the above criteria is equivalent to the VAE criteria in Eq. (\ref{equ:vae_family})
\[ \mathcal{L}_{VAE} = E_{p_{data}(x)}E_{q_\phi(z|x)}[ \log p_\theta(x|z)] - R(q_\phi) + C \]
where $C$ is a normalization constant irrelevant to the optimization. This means that the family $\mathcal{P}$ that we have chosen is actually the family of fixed variance factored Gaussians $\mathcal{P} = \lbrace \mathcal{N}(\mu, I/2) | \mu \in \mathbb{R}^N \rbrace$. According to Proposition 2, this objective will attempt to approximate the posterior $q_\phi(x|z)$, the distribution over data points that map to $z$, with a fixed variance Gaussian.
Unfortunately, common distributions such as natural images almost never have a mixture of Gaussian structure: if $x$ is a likely sample, $x + \mathcal{N}(0, I/2)$ is not. 
Unless $q_\phi(z|x)$ is lossless, it will map multiple $x$ to the same encoding $z$, resulting in a highly non-Gaussian posterior $q_\phi(x|z)$.
This is where the fuzziness comes from: 
the mean of the best fitting Gaussian is some "average" of $q_\phi(x|z)$. 
Formally we have the following proposition.
\begin{prop}
\label{prop:reconstruction_error}
The optimal solution to reconstruction loss of Eq.(\ref{equ:reconstruction_loss}) for a given $q_\phi$ is 
\[ g_\theta(z) = \E_{q_\phi(x|z)} [x]
\]
and the optimal expected reconstruction error $E_{q_\phi(x|z)}[||g_\theta(z) - x||_2^2]$ for any $z$ is the sum of coordinate-wise variances $\sum_{i} Var_{q_\phi(x|z)}[x_i]$.
\end{prop}

Intuitively Proposition \ref{prop:reconstruction_error} follows from the observation that the optimal $p_\theta(x|z)$ is an M-projection onto $q_\phi(x|z)$, and therefore satisfies moment matching conditions. It shows that the optimal reconstruction is an average of $q_\phi(x|z)$ and $\sum_{i} Var_{q_\phi(x|z)}[x_i]$ measures the reconstruction error.
For image data this error is reflected by blurry samples. 

\begin{figure}
\centering
\sbox0{\includegraphics{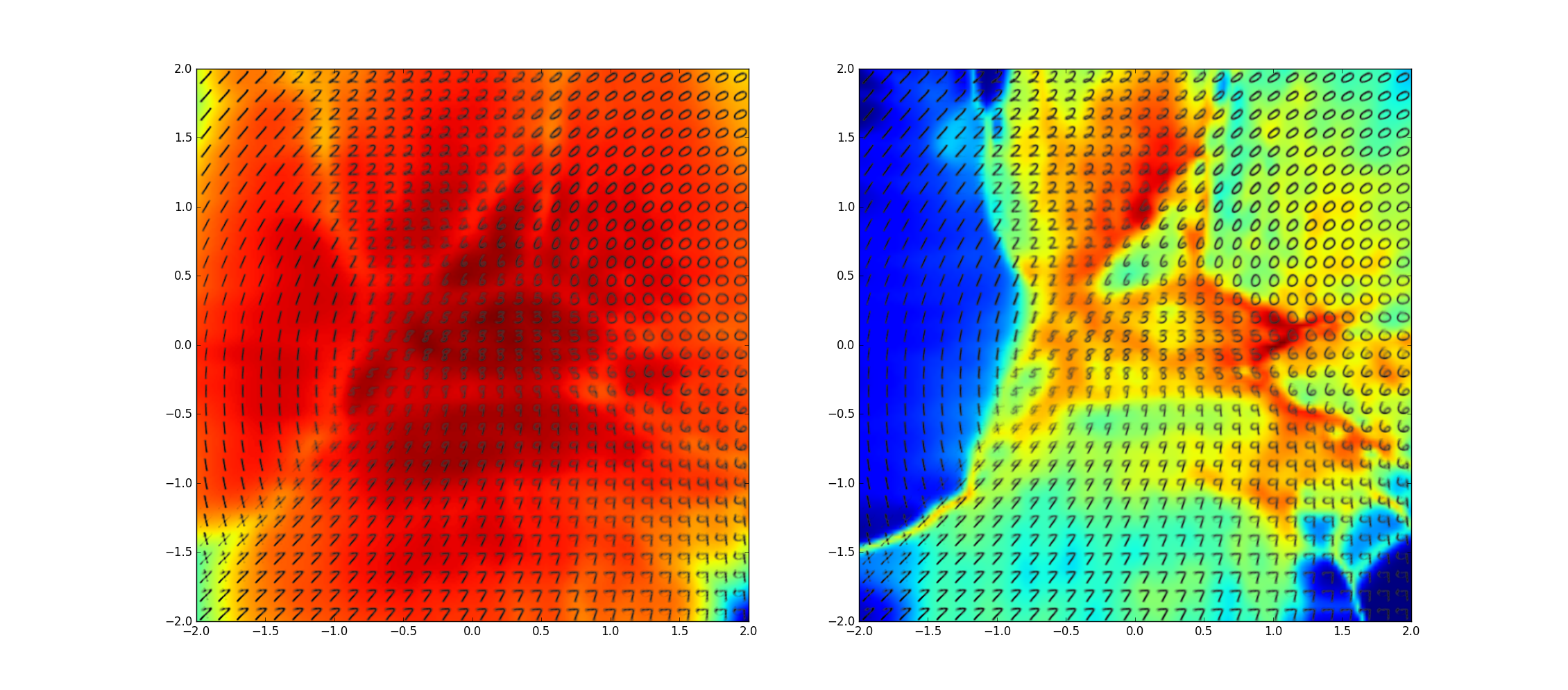}}%
\includegraphics[clip,trim={.5\wd0} 30 70 30,width=\linewidth]{plot/coverage_vs_ambiguity}
\caption{$\sum_{i} Var_{q_\phi(x|z)}[x_i]$ plotted on latent space, red corresponds to high variance, and blue low variance. Plotted digits are the generated $g_\theta(z)$ at any $z$. Digits on high variance regions are fuzzy while digits on low variance regions are generally well generated. (Best viewed on screen)}
\label{fig:coverage_and_ambiguity}
\end{figure}

We illustrate this fact by fitting a VAE on MNIST with 2 dimensional latent code using the ELBO regularization (\ref{equ:elbo_criteria}) and 2-norm (Gaussian) reconstruction loss (\ref{equ:reconstruction_loss}). In Figure \ref{fig:coverage_and_ambiguity} we plot for each $z \in \mathcal{Z}$ the posterior variance $\sum_{i} Var_{q_\phi(x|z)}[x_i]$ (color coded) and the digits generated by $g_{\theta^*}(z)$.  
Regions of latent space $\mathcal{Z}$ where $q_\phi(x|z)$ has high variance (red) correspond to regions where "fuzzy" digits are generated. 

The problem of fuzzy samples in VAEs was previously attributed to the maximum likelihood objective
(which penalizes regions where $p_{data}(x) \gg p_\theta(x)$ more than regions where $p_{data}(x) \ll p_\theta(x)$), thus encouraging solutions $p_\theta(x)$ with larger support.
This explanation was put into question by \citep{f_gan2016}, who 
showed that no major difference is observed when we optimize over different types of divergences with adversarial learning. 
Our conclusion is consistent with this recent observation, in that fuzziness is not a direct consequence of maximum likelihood, but rather, a consequence of the VAE approximation of maximum likelihood. 


We will show similar results for other distribution families $\mathcal{P}$ in the Appendix.



\subsection{Infusion Training as Latent Code Augmentation}
The key problem we observed in the previous section is that an insufficiently discriminative (mapping different $x$ to the same $z$) feature detector $q$ will have a posterior too complex to be approximated by a simple family $\mathcal{P}$. In this section we propose a method to alleviate this problem and achieve significantly sharper samples on complex natural image datasets. In particular, we draw a connection with and generalize the recently proposed infusion training method~ \citep{infusion_training2016}.

\begin{figure}
\centering
\begin{subfigure}[t]{0.48\linewidth}
\centering

\includegraphics[width=\textwidth]{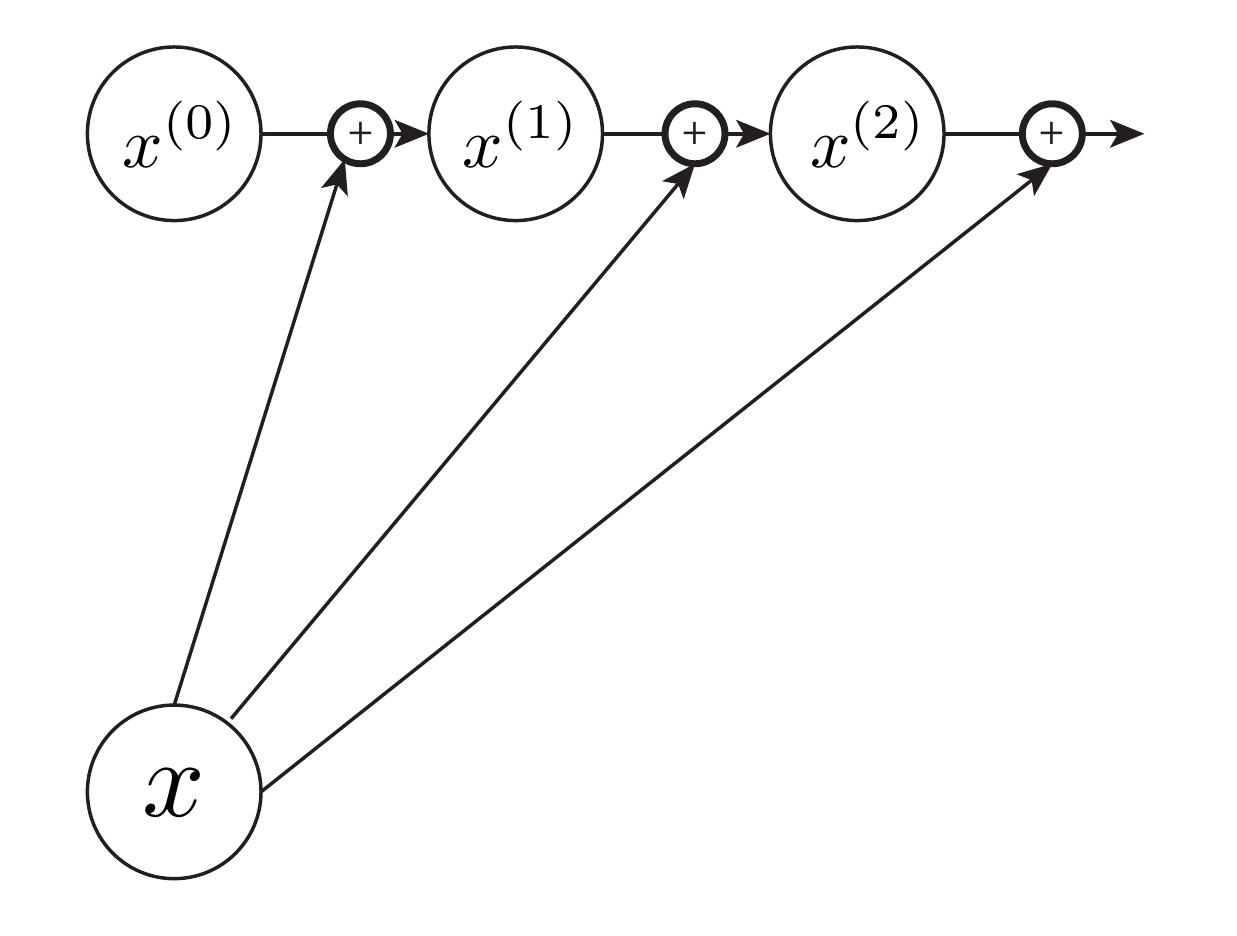}
  



\end{subfigure}
~
\begin{subfigure}[t]{0.48\linewidth}
\centering
\includegraphics[width=\textwidth]{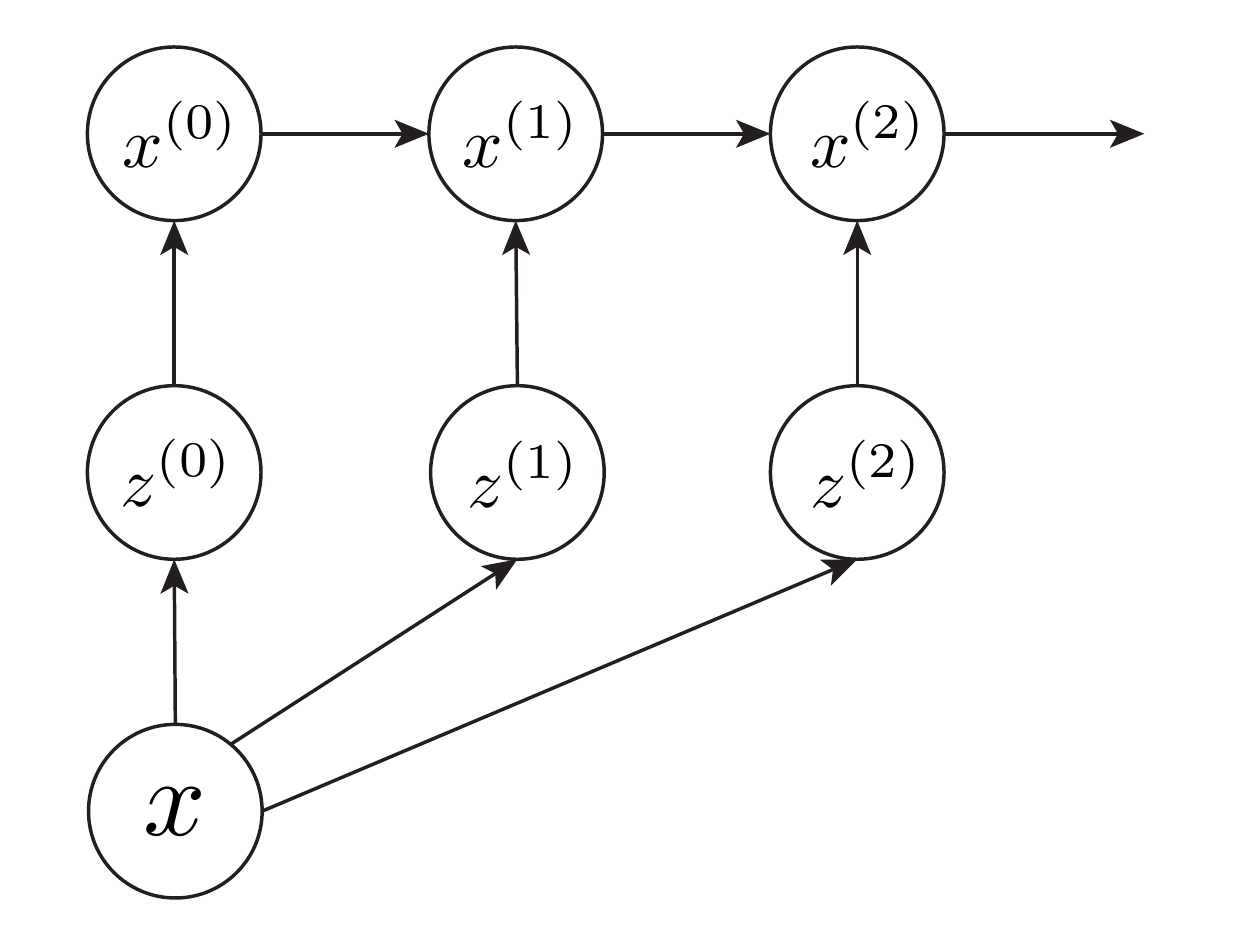}



\end{subfigure}
\caption{Infusion Training (Left) vs. Sequential VAE (Right). For Infusion Training, at each step some random pixels from real data are added to the previous reconstruction. Based on the newly added pixels the model makes a new attempt at reconstruction. Sequential VAE is a generalization of this idea. At each step some features are extracted from real data. The network makes a new attempt at reconstruction based on previous results and the new information. }
\label{fig:infusion_illustration}
\end{figure}

\begin{figure*}[h]
    \centering
    \begin{subfigure}[t]{0.4\textwidth}
        \centering
        \includegraphics[width=\textwidth]{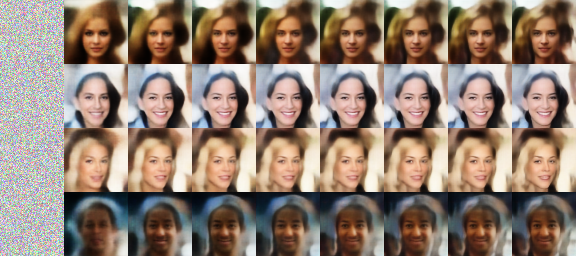}
    \end{subfigure}%
    ~ 
    \begin{subfigure}[t]{0.4\textwidth}
        \centering
        \includegraphics[width=\textwidth]{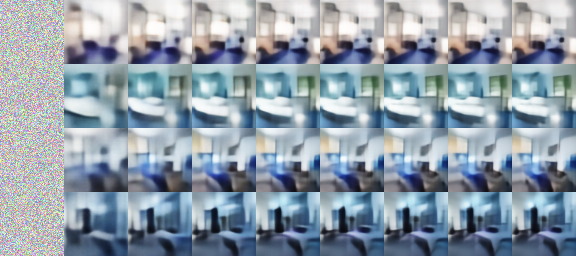}
    \end{subfigure}
    \caption{Sequential VAE on CelebA and LSUN. Each column corresponds to a step in the sequence (starting from noise); in particular, the second is what a regular VAE with the same architecture generates. We see increasingly sharp images and addition of details with more iterations (from left to right).}
    \label{fig:infusion_vae_experiments}
\end{figure*}

Infusion training \citep{infusion_training2016} trains a Markov chain to gradually converge to the data distribution $p_{data}$. Formally, training 
starts with some initial random noise $x^{(0)} \sim p^{(0)}$, and goes through the following two steps iteratively

\textbf{1)Infusion:} A new "latent state" $z^{(t)}$ is generated by taking the previous reconstruction $x^{(t)}$, and adding some pixels from a ground truth data point $x_{data}$. 

\textbf{2)Reconstruction:} The decoding model attempts the next reconstruction $p^{(t)}(x^{(t+1)}|z^{(t)})$ by maximizing $ \log p^{(t)}(x_{data}|z^{(t)}) $. The superscript $t$ indicates that this can be a different distribution for each step $t$, leading to a non-homogeneous Markov chain.

To draw new samples at test time, we directly sample from the Markov chain
\[ p^{(t)}(x^{(t)}|x^{(t-1)}) \] 
initializing from random noise.
This idea is illustrated in Figure~\ref{fig:infusion_illustration}. 
Note that we refer to the resulting image after infusion $z^{(t)}$ as a "latent state" because it plays the same role as a VAE latent state. We can interpret the probability of obtaining $z^{(t)}$ by the above iterative procedure 
as an {\em inference} distribution $q(z^{(t)}|x)$. 
In contrast with VAEs, the inference distribution used in infusion training is manually specified by the "infusion" process.
By adding more true pixels and making $z^{(t)}$ increasingly informative about $x$, for sufficiently large $t$ the "latent code" $z^{(t)}$ will become informative enough to have a simple posterior $q(x|z^{(t)})$ that is highly concentrated on $x$. Such a posterior can be well approximated by simple unimodal conditionals $\mathcal{P}$, such as Gaussian distributions. 

Inspired by this idea, we propose the model shown in Figure~\ref{fig:infusion_illustration} which we will call a \textbf{sequential VAE}. Each step is a VAE, except the decoder $p_\theta(x|z)$ is now also conditioned on the previous reconstruction outcome. We go through the following two steps iteratively during training:

\textbf{1) Inference:} An inference distribution $q_{\phi_t}(z^{(t)}|x)$ maps a ground truth data point $x_{data}$ to a latent code. 

\textbf{2) Reconstruction:} A generative distribution (decoder) that takes as input a sample from the previous step $x^{(t-1)}$ and latent code $z^{(t)}$ to generate a new sample $p_{\theta_t}(x^{(t)}|z^{(t)}, x^{(t-1)})$. When $t=0$, we do not condition on previous samples. 

The model is jointly trained by maximizing the VAE criteria for each time step respectively.
\[ \log p_{\theta_t}(x_{data}|z^{(t)}, x^{(t-1)}) + R(q_{\phi_t}) \]
 For experiments in this section we use ELBO regularization \citep{autoencoding_variational_bayes2013} $R(q_{\phi_t}) = KL(q_{\phi_t}(z^{(t)}|x_{data})||p(z^{(t)}))$ where $p(z^{(t)})$ is a simple fixed prior such as white Gaussian.

To generate samples during test time, for each step we perform ancestral sampling $p(z^{(t)})p_{\theta_t}(x^{(t)}|z^{(t)}, x^{(t-1)})$. Details about implementation is described in the Appendix.

The idea is that the more latent code we add, the more we know about $x$, making the posterior $q(x|z^{(0:t)})$ simpler as $t$ becomes larger. In particular, we can show this formally for 2-norm loss as in Section~\ref{sec:gaussian_limitation}.

\begin{prop}
\label{prop:sequential_vae}
For any distribution $q$, and any $z^{(0:t-1)}$, and input dimension $i$,
\[ Var_{q(x|z^{(0:t-1)})}[x_i] \geq E_{q(z^{(t)})}\left[ Var_{q(x|z^{(0:t)})}[x_i] \right] \]
\end{prop}
Therefore increasing the latent code size in expectation does not increase variance. By the connection we established between variance of the posterior and blurriness of the samples, this should lead to sharper samples. We show this experimentally in Figure~\ref{fig:infusion_vae_experiments}\footnote{Code is available at https://github.com/ShengjiaZhao/Sequen\\tial-Variational-Autoencoder}, where we evaluate our model on CelebA and LSUN. In particular we can generate sharp LSUN images based only on 2-norm loss in pixel space, something previously considered to be difficult for VAE models. Details about architecture and training are in the Appendix. 

Sequential generation is a general scheme under which many different models are possible. 
It encompasses infusion training as a special case,
but many different variants are possible. This idea has great potential for improving auto-encoding models based on simple, unimodal $\mathcal{P}$ families. 

\section{Complex $\mathcal{P}$ and the Information Preference Property}
\label{sec:information_preference}

\begin{figure*}[t!]
\centering
\begin{tabular}{ccc}
\includegraphics[height=0.15\linewidth]{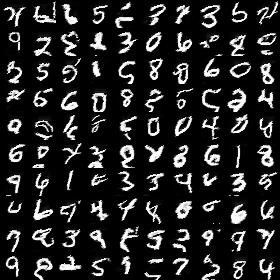} & 
\includegraphics[height=0.15\linewidth]{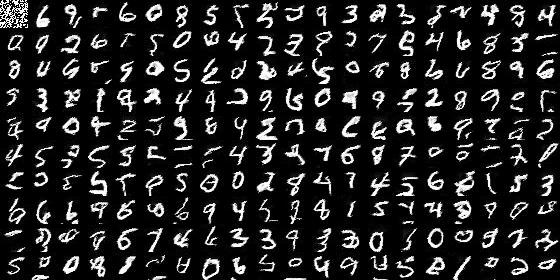} & 
\includegraphics[height=0.15\linewidth]{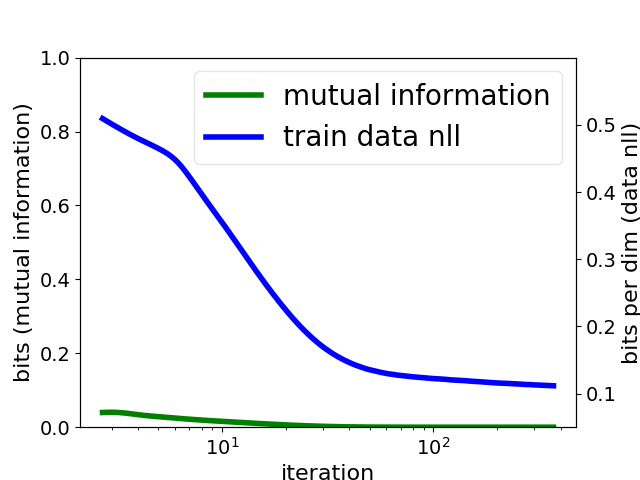} \\ 
\includegraphics[height=0.15\linewidth]{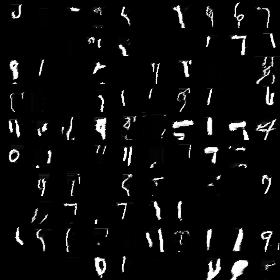} &
\includegraphics[height=0.15\linewidth]{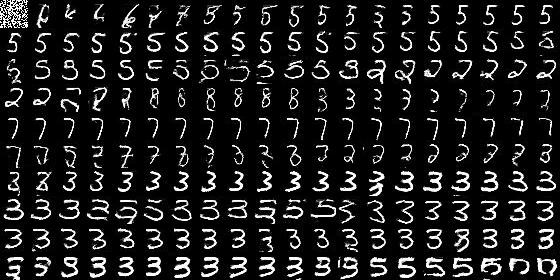} & 
\includegraphics[height=0.15\linewidth]{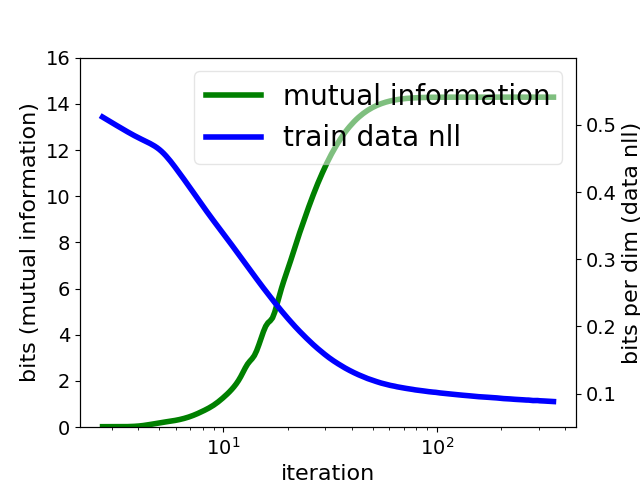}
\end{tabular}
\caption{Mutual information vs sample quality for VAE with PixelCNN as family $\mathcal{P}$. Top row: Pixel VAE optimized on ELBO bound. Bottom row: Pixel VAE optimized without regularization. For ELBO ancestral sampling (Left) $p(z)p_\theta(x|z)$ produces similar quality samples as Markov chain (Middle), while for unregularized VAE ancestral sampling produces unsensible samples, while Markov chain produces samples of similar quality as ELBO. Right: evolution of estimated mutual information and per-pixel negative log likelihood loss. For ELBO, mutual information is driven to zero, indicating unused latent code, while without regularization large mutual information is preferred. Details on the mutual information approximation is in the Appendix.
}
\label{fig:pixel_vae_mnist}
\end{figure*}

\begin{figure*}[h]
\centering
\begin{tabular}{ccc}
\includegraphics[height=0.15\linewidth]{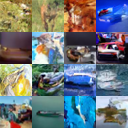} &
\includegraphics[height=0.15\linewidth]{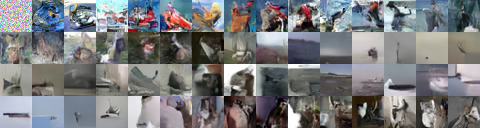} & 
\includegraphics[height=0.15\linewidth]{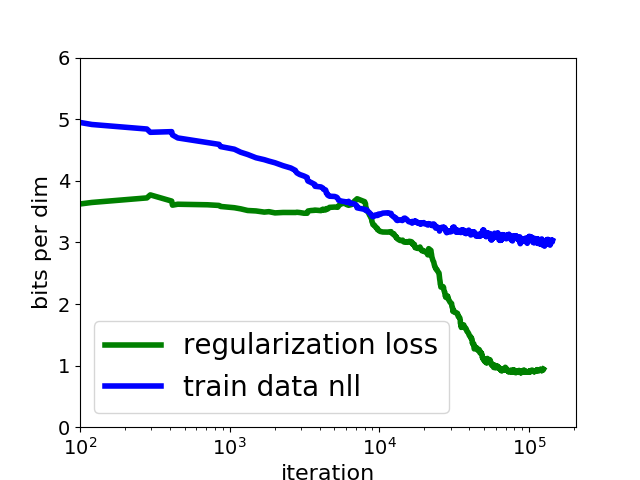} \\
\includegraphics[height=0.15\linewidth]{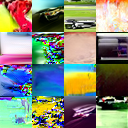} &
\includegraphics[height=0.15\linewidth]{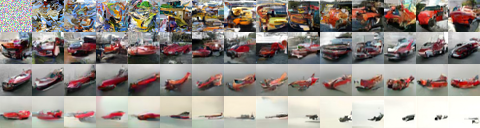} & 
\includegraphics[height=0.15\linewidth]{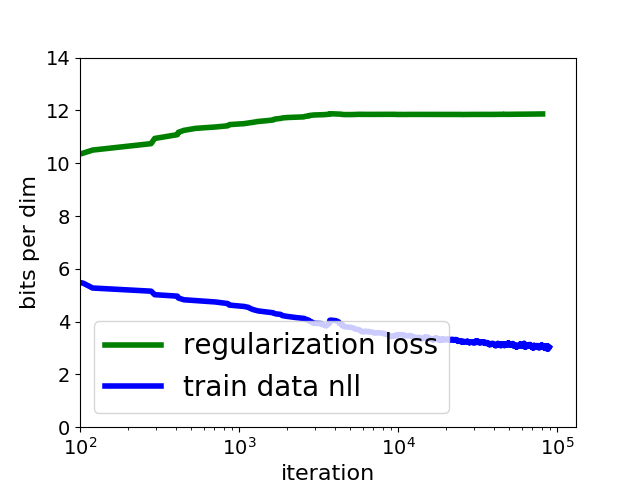}
\end{tabular}
\caption{Experiment on CIFAR with PixelCNN as family $\mathcal{P}$. Meaning of plots is identical to Figure~\ref{fig:pixel_vae_mnist}. The only difference is that CIFAR is too complex for our PixelCNN model to directly model, so the latent code is used in both cases. In both cases the mutual information is too 
difficult to directly estimate. Therefore we plot $KL(q(z|x)||p(z))$ instead.}
\label{fig:pixel_vae_cifar}
\end{figure*}

Models that use a complex $\mathcal{P}$ such as recurrent density estimators have demonstrated good promise in modeling complex natural datasets \citep{pixel_vae2016}. However these models have a shortcoming demonstrated in \citep{lossy_vae2016}. A model with a complex conditional distribution and optimized under the ELBO criterion tend to ignore the latent code. \citep{lossy_vae2016} gave an explanation of this information preference property using coding theory. Here we provide an alternative simple explanation using the framework introduced int this paper . A equivalent way to write the ELBO criteria Eq.(\ref{equ:elbo_criteria}) is as the negative sum of two divergences \citep{autoencoding_variational_bayes2013}
\begin{align*}
\mathcal{L}_{ELBO} &= -KL(p_{data}(x)||p_\theta(x)) - \\ 
& \qquad E_{p_{data}(x)}[KL(q_\phi(z|x)||p_\theta(z|x))] \leq 0 \numberthis \label{equ:elbo_information_preference}
\end{align*}
Suppose $\mathcal{P}$ is sufficiently large, so that there is a member $p^* \in \mathcal{P}$ that already satisfies $KL(p_{data}(x)||p^*(x)) = 0$. If the second divergence is also $0$, then this is already the best we can achieve. The model can trivially make the second divergence $0$ by making latent code $z$ completely non-informative, i.e., making $z$ and $x$ independent under both $p_\theta$ and $q_\phi$, so that $p_\theta(z|x) = p(z)$, $q_\phi(z|x) = p(z)$. There is no motivation for the model to learn otherwise, undermining our purpose of learning a latent variable model.

However this problem can be fixed using the general VAE objective we introduced in Eq.(\ref{equ:vae_family})
\[ \mathcal{L}_{VAE} = \E_{q_\phi(x, z)}[\log p_\theta(x|z)] - R(q_\phi) \]
If we do not regularize (and therefore do not attempt to meet the conditions in Proposition~\ref{prop:marginal_condition}, setting $R(q_\phi)=0$, there is an incentive to use the latent code. This is because if we satisfy conditions in Proposition~\ref{prop:condition}, we have
\begin{align*}
\E_{q_\phi(x, z)}[\log p_{\theta^*}(x|z)] &= \E_{q_\phi(z)}\E_{q_\phi(x|z)} [\log q_\phi(x|z)] \\
&= \E_{q_\phi(z)}[-H_{q_\phi}(x|z)] \\
&= I_{q_\phi}(x; z) - H_{p_{data}}(x)
\end{align*}
where $H_q$ is the entropy under some distribution $q$, and $I_q$ the mutual information. This means that this optimization criteria \textbf{actually prefers to maximize mutual information between $x$ and $z$ under $q$}, unlike the ELBO objective (\ref{equ:elbo_information_preference}).

We have derived before that $R(q_\phi)$ is not needed if we do not require sampling to be tractable 
, as we will still be able to sample by running a Markov chain as in Proposition~\ref{prop:condition}. If the goal is to encode the data distribution and learn informative features, then we can ignore $R(q_\phi)$ and the objective will encourage the use of the latent code. We illustrate this on a model that uses PixelCNN \citep{pixelcnn_pp2017, pixel_rnn2015, conditional_pixelcnn2016, pixel_vae2016} as the family $\mathcal{P}$. The results are shown in Figure~\ref{fig:pixel_vae_mnist} and Figure~\ref{fig:pixel_vae_cifar}. Experimental setting is explained in the appendix.\footnote{Code is available at https://github.com/ShengjiaZhao/Genera\\lized-PixelVAE}

On both MNIST and CIFAR, we can generate high quality samples with or without regularization with a Markov chain. As expected, only regularized VAE produces high quality samples with ancestral sampling $p(z)p_\theta(x|z)$, as it encourages satisfaction of the condition in Proposition~\ref{prop:marginal_condition}. However, mutual information $I_q(x; z)$ between data and latent code is minimized with the ELBO criterion as shown in the top right plot in Figure~\ref{fig:pixel_vae_mnist} and \ref{fig:pixel_vae_cifar}. In fact, mutual information is driven to zero in Figure~\ref{fig:pixel_vae_mnist}, indicating that the latent code is completely ignored. On the other hand, for unregularized VAE high mutual information is preferred as shown in the bottom right plot of Figure~\ref{fig:pixel_vae_mnist} and \ref{fig:pixel_vae_cifar}. 



\section{Conclusion}
In this paper we derived a general family of VAE methods from a new perspective, which is not based on lower bounding the intractable marginal likelihood. Instead, we take the perspective of a variational approximation of the posterior of an inference distribution or feature detector. Using this new framework, we were able to explain some of the issues encountered with VAEs: blurry samples and the tendency to ignore the latent code. Using the insights derived from our new framework, we identified two new VAE models that singnificantly alleviate these problems.

\FloatBarrier

\section{Acknowledgements}
We thank Justin Gottschlich, Aditya Grover, Volodymyr Kuleshow and Yang Song for comments and discussions. This research was supported by Intel, NSF (\#1649208) and Future of Life Institute (\#2016-158687). 

\bibliographystyle{icml2017}
\begin{small}
\bibliography{ref}
\end{small}

\newpage
\FloatBarrier
\newpage
\appendix
\section{Additional Results}
\subsection{Comparison to Adversarial Training}
Adversarial training \citep{generative_adversarial_nets2014} has shown great promise in generating high quality samples. Analysis in this paper points to a possible explanation of its relative success in complex natural image datasets compared to variational autoencoders. In Proposition~\ref{prop:optimal_solution} we pointed out that if $q(x|z_1) = q(x|z_2)$ then the model has no incentive to take advantage of this representation capacity by mapping them to different members of $\mathcal{P}$. This is the key reason why a simple family $\mathcal{P}$ require a almost "lossless" $q$, and failure to satisfy this condition leads to fuzziness and other problems. 

However adversarial training does not suffer from this limitation. In fact even without inference, adversarial training can map different latent code $z_1, z_2$ to different members of $\mathcal{P}$. This is because for adversarial training we are not using $\mathcal{P}$ to approximate $q(x|z)$. Instead we are selecting a member of $\mathcal{P}$ whose support is covered by the support of real data. Intuitively, we would like to generate {\em any} real looking samples, and not a particular set.

Therefore we expect adversarial training to have advantage over VAE when $q(x|z)$ is expected to be complex, and $\mathcal{P}$ is simple. However we showed in this paper that models with complex $\mathcal{P}$, but carefully designed to avoid the information preference property also show great promise. 

\subsection{Failure Modes for Factorized Discrete Family}
\stefano{this subsection can potentially go in appendix}
\label{sec:vae_norm_loss}
When $\mathcal{P}$ is the family of factorized discrete distribution $p(x) = \prod_{i=1}^{N} p_i(x)$, where each $p_i$ is a discrete distribution on the $i$-th dimension. We obtain similar results

\begin{prop}
\label{prop:discrete_optimum}
The optimal solution $\theta^*$ to $\mathcal{L}_{VAE}$ when $\mathcal{P}$ is the family of discrete distribution for a given $q_\phi$ is for all $i$, $z$
\[ p_{\theta^*}(x_i|z) = q_\phi(x_i|z) \]
and for each $z$ the best achievable error $E_{q_\phi(x|z)}[\log p_\theta(x|z)]$ is the pixel-wise negative entropy $\sum_i H(q_\phi(x_i|z))$
\end{prop}

This shows us that for discrete distributions, mismatch between $q_\phi$ and $\mathcal{P}$ manifests in a different way: by generating excessively noisy output where each pixel is independently sampled. \shengjia{Add similar plot as Gaussian}

\subsection{Estimating Mutual Information}
Because
\[ I_q(x, z) = H_q(z) - H_q(x|z) = E_{q(x, z)}\left[ \log \frac{q(z|x)}{q(z)} \right] \]
We can estimate mutual information by obtaining $M$ samples $x_i, z_i \sim q(x, z)$, and 
\begin{align*}
\tilde{I}_q(x, z) \approx \frac{1}{M} \sum_i \left[ \log \frac{q(z_i|x_i)}{1/M \sum_j q(z_i|x_j)} \right]
\end{align*}
This gives us good estimates unless the mutual information is large because the above estimation is upper bounded by $\log M$
\[ \tilde{I}_q(x, z) \leq \log M \]
This problem is not specific to our method of approximation. In fact, suppose the dataset has $M$ samples, then \textit{true} mutual information under the empirical data distribution is also upper bounded by
\[ I(x, z) = H(x) - H(x|z) \leq \log M \]

\section{Proofs}
\begin{proof}[Proof of Proposition~\ref{prop:optimal_solution}\ref{prop:condition}\ref{prop:marginal_condition}]
For any $z$, because  
\[ E_{q(x|z)} [\log p_\theta(x|z)] \leq \max_{p \in \mathcal{P}} E_{q(x|z)}[\log p(x)] \]
When $\mathcal{F}$ is a sufficiently large family, there must be a $f^* \in \mathcal{F}$ so that for all $z \in \mathcal{Z}$, $f^*(z) \in \arg\max_{p \in \mathcal{P}} E_{q(x|z)}[\log p(x)]$. Therefore 
\begin{align*} 
\mathcal{L} &\leq E_{q(z)}\left[\max_{p \in \mathcal{P}} E_{q(x|z)}[\log p(x)]\right] \\ 
&= E_{q(z)} E_{q(x|z)} [\log f^*(z)(x)] \numberthis \label{equ:optimal_bound} 
\end{align*}
which means that $f^*$ is the global maximum of $\mathcal{L}$. Note that for any distribution $p$, we have
\[ E_{q(x|z)}[\log p(x)] \leq E_{q(x|z)}[\log q(x|z)] \]
from the non-negativity property of KL-divergence. If condition 2 is satisfied, i.e.,  $\forall z \in \mathcal{Z}$, $q(x|z) \in \mathcal{P}$, then the optimum in Equation (\ref{equ:optimal_bound}) is attained by
\[ f^*(z) = q(\cdot|z) \]
for all $z \in \mathcal{Z}$. Then 
\[ q(z)f^*(z)(x) = q(z) q(x|z) = q(x, z) \]
which by definition has marginal $p_{data}(x)$. Finally if condition 3 is satisfied, then 
\[ p(z)f^*(z)(x) = q(z) f^*(z)(x) = q(x, z) \]
which also has marginal $p_{data}(x)$.
\end{proof}

\begin{proof}[Proof of Proposition~\ref{prop:reconstruction_error}]
Given $z$ the $x^*$ that maximizes $E_{q(z|x)}[||x^* - x||_2^2]$ is given by
\begin{align*}
\nabla_{x^*} E_{q(x|z)}[||x^* - x||_2^2] &= 0 \\
E_{q(x|z)}[x^* - x] &= 0 \\
x^* = E_{q(x|z)}[x] &= \mu[q(x|z)]
\end{align*}
That is the optimal $x^*$ is simply the mean of $q(x|z)$, and under this $x^*$, $E_{q(z|x)}[||x^* - x||_2^2] = Var[q(x|z)]$. The optimal $f^*$ must also map each $z$ to this $x^*$ because
\begin{align*}
\mathcal{L}_{VAE} &= E_{p_{data}(x)}E_{q(z|x)}[||f(z) - x||_2^2] - R(q) \\
&\leq E_{q(z)}\left[\max_{\hat{x}} E_{q(x|z)}[||\hat{x} - x||_2^2]\right] - R(q) \\
&= E_{q(z)}\left[E_{q(x|z)}[||f^*(z) - x||_2^2]\right] - R(q)
\end{align*}
\end{proof}

\begin{proof}[Proof of Proposition~\ref{prop:discrete_optimum}]
Because $\mathcal{P}$ is the family of factorized discrete distribution, denote each member of $\mathcal{P}$ as $p = (p_1, \cdots, p_N)$ where $p_i$ is the independent probability of the $i$-th dimension taking value $1$ instead of $0$, the loss for each $z$ can be written as
\begin{align*}
E_{q(x|z)} & \left[ \log \prod_{i=1}^{N} p_i^{x_i} (1 - p_i)^{1 - x_i} \right] \\ 
	&= \sum_{i=1}^{N} E_{q(x_i|z)}[x_i \log p_i + (1 - x_i) \log (1 - p_i)] 
\end{align*}
and the optimal solution to the above satisfies
\begin{align*}
\nabla_{p_i} E_{q(x|z)} \left[ \log \prod_{i=1}^{N} p_i^{x_i} (1 - p_i)^{1 - x_i} \right] = 0  \\
\end{align*}
whose unique solution is $p_i = q(x_i|z)$. We can further compute that the optimal loss as
\begin{align*}
\sum_{i=1}^{N} E_{q(x_i|z)} & [p_i \log p_i + (1 - p_i) \log (1 - p_i)] \\ 
&= \sum_{i=1}^N H(q(x_i|z)) 
\end{align*}
\end{proof}



\section{Experimental Setup}
\subsection{Sequential VAE}
\begin{figure}
\centering
\includegraphics[width=\linewidth]{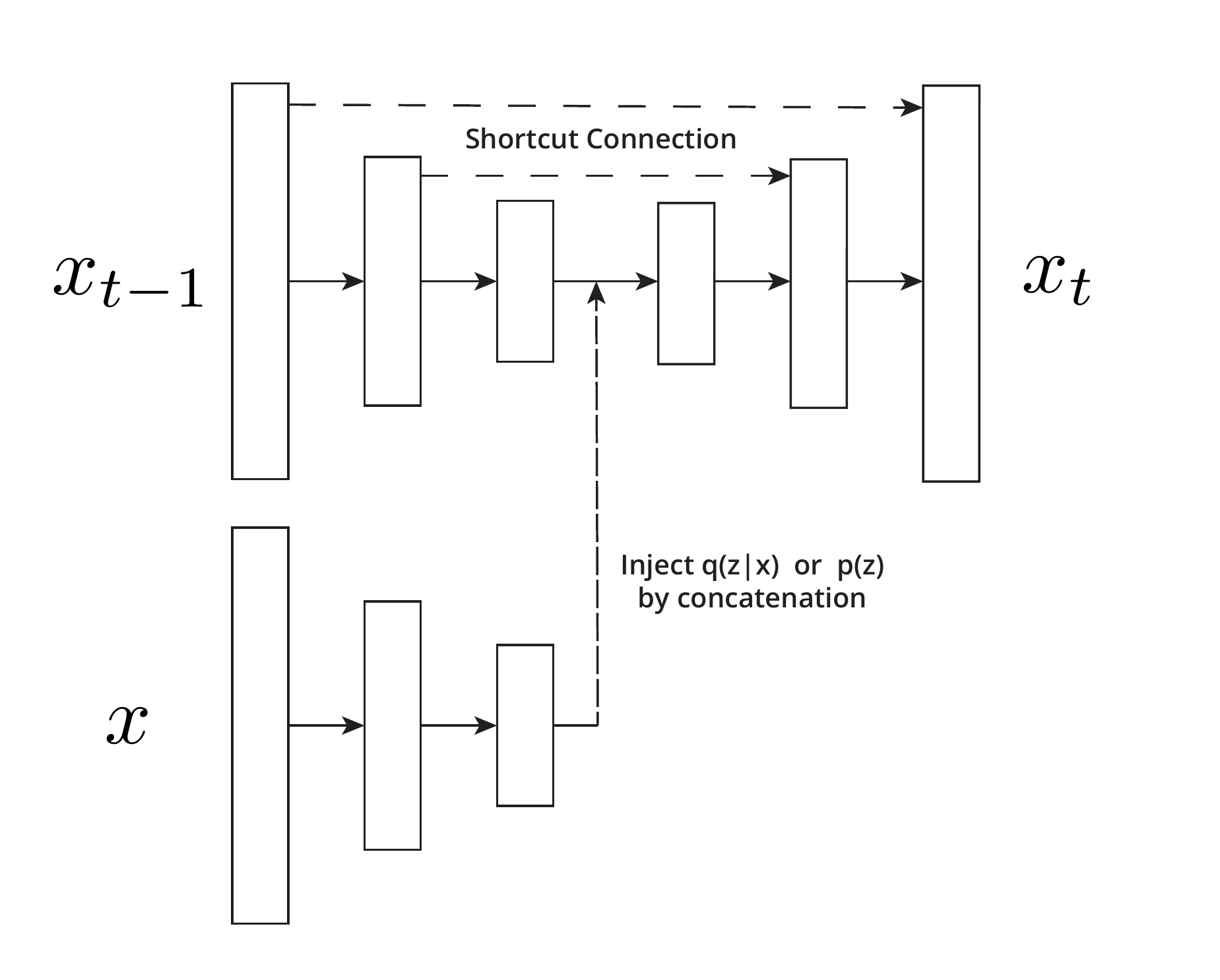}
\caption{Architecture for Sequential VAE. Each rectangle in the figure represents a few convolution steps followed by upsampling or downsampling. In the experiments, two convolution layers are used before each up or down sample}
\label{fig:seq_vae_architecture}
\end{figure}

Each step of the Sequential VAE is contains an encoder that takes as input a ground truth $x$ and produces latent code $q^{(t)}(z^{(t)}|x)$, and an autoencoder with short cut connections which takes as input the output from previous step $x^{(t-1)}$, and latent code $z^{(t)}$ that is either generated from prior (during testing) or by encoder (during training). Short cut connection encourages the learning of identity mapping to help the model preserve and refine upon the results from previous step. This is either achieved by direct addition or by gated addition with learnable parameter $\alpha$
\[ \hat{z} = \alpha \cdot z + (1 - \alpha) z_{new} \]
The architecture is shown in Figure~\ref{fig:seq_vae_architecture}. We use a non-homogeneous Markov chain so weights are not shared between different time steps. For detailed information about the architecture please refer to https://github.com/ShengjiaZhao/Sequential-Variational-Autoencoder

\subsection{VAE with PixelCNN}
For MNIST we use a simplified version of the conditional PixelCNN architecture \citep{conditional_pixelcnn2016}. For CIFAR we use the public implementation of PixelCNN++ \citep{pixelcnn_pp2017}. In either case we use a convolutional network to generate a 20 dimensional latent code, and plug this into the conditional input for both models. The entire model is trained end to end with or without regularization. For detailed information please refer to https://github.com/ShengjiaZhao/Generalized-PixelVAE.

\end{document}